\documentclass[conference]{IEEEtran}
\usepackage[utf8]{inputenc}
\usepackage{graphicx}
\usepackage{color}
\usepackage{amsmath,amssymb}
\usepackage{subfigure}
\usepackage[standard]{ntheorem}
\usepackage{url}

\title{Finding optimal finite biological sequences over finite alphabets: the OptiFin toolbox}

\author{
\IEEEauthorblockN{R\'egis Garnier, Christophe Guyeux, and St\'ephane Chr\'etien}
\IEEEauthorblockA{Femto-ST Institute, UMR 6174 CNRS
\& Laboratory of Mathematics, Besan\c con\\
Université de Bourgogne Franche-Comté, France\\
Email: christophe.guyeux@univ-fcomte.fr}}

\begin{document}

\maketitle

\begin{abstract}
In this paper, we present a toolbox for a specific optimization problem that frequently arises in bioinformatics or genomics. In this specific optimisation problem, the state space is a set of words of specified length over a finite alphabet. To each word is associated a score. The overall objective is to find the words which have the lowest possible score. This type of general optimization problem is encountered in e.g 3D conformation optimisation for protein structure prediction, or largest core genes subset discovery based on best supported phylogenetic tree for a set of species. In order to solve this problem, we propose a toolbox that can be easily launched using MPI and embeds 3 well-known metaheuristics. The toolbox is fully parametrized and well documented. It has been specifically designed to be easy modified and possibly improved by the user depending on the application, and does not require to be a computer scientist. We show that the toolbox performs very well on two difficult practical problems.
\end{abstract}

\section{Introduction}

Many biological data can be described as finite sequences over a finite alphabet. For instance, from a computer science viewpoint, a gene is nothing but a sequence of nucleotides $A$, $C$, $G$, $T$, and a protein is a sequence of amino acids (a.a., a set of 20 elements). In various situations, we can assign a score value to each finite sequence, that can account for its ability to solve a biological problem. For instance, these finite sequences can be the mapped to the descriptions of the protein's spatial configuration in the 3D square lattice; in this case, the score of the considered conformation is simply its energy. The main problem considered in this paper is the one of optimising the score value over all possible sequences. As is well known, most such problems are NP-hard and we often have to resort to efficient heuristics. 

The main contribution of the present paper is to propose an handy and efficient toolbox that can solve such problems using a combination of metaheuristics. The toolbox is carefully designed in such a way that the sequences, their updates and the score function are easy to define and work with. Our toolbox -- which is still in alpha stage, but in active development -- incorporates the state of the art computation standards in computer science. It is mainly written in Python, it is MPI compliant, and it uses XML configuration files. The toolbox can be launched on a collection of computers through the network. After completion, the toolbox provides a report summarising various information concerning its convergence. We provide examples of application to phylogenetics and protein structure predictions.

The remainder of this article is as follows. In the next section, we present the main problem and motivate its relevance to bioinformatics via some illustrative examples. In Section~\ref{sec:3heuristics}, 3 currently embedded metaheuristics are presented. The toolbox's architecture is described in Section~\ref{sec:software}. Application examples are provided in Section~\ref{sec:appli}. The article ends with a conclusion, in which the contribution is summarized and future work is outlined.

\section{What types of problems ?}

\subsection{Main objective}
\label{subsec:main}
Given an integer $n>0$ and a finite set $\mathcal{A}$, we consider the set $\mathfrak{S}_n$ of sequences of length $n$ whose elements belong to $\mathcal{A}$. In an equivalent manner, $\mathfrak{S}_n$ is the finite set of $n$-words $w=w_1...w_n$ whose letters $w_i$ belongs to the alphabet $\mathcal{A}$. We consider a distance $d:\mathfrak{S}_n \times \mathfrak{S}_n \longrightarrow \mathbb{R}^+$, and we suppose that $d\left(\mathfrak{S}_n, \mathfrak{S}_n\right) \subset \mathbb{N}^\ast$: all distances between words are integer values. Finally, we consider a ``scoring function'' $s:\mathfrak{S}_n \longrightarrow \mathbb{R}^+$, that is, a function that maps each $n$-word to a real positive value. Our objective is to design a software that uses various heuristics to find the minimum value of $s$ on $\mathfrak{S}_n$.

Another way of seeing is to construct an undirected graph. Each $n$-word can be identified with a node of a graph, while two nodes $w$ and $w'$ are connected with an edge if and only if $d(w,w')=1$. At each node is associated its score. Our goal is then to design a way of moving from a node to the next in the graph, in such a way that the scoring function decreases over successive iterations. As will be clarified later, some sort of continuity will be needed : two nodes close enough must have, most of the times, approximately the same score. Some jumps will be acceptable, but should remain quite rare.

We will now motivate that this kind of problem arises in various instances in the field of bioinformatics by providing three important applications.

\subsection{A first example: core genes and phylogeny}
In the first example, we consider a set of species from which we want to infer a biomolecular phylogeny. For each species, we have a list of genes, i.e. a dictionary whose key is the gene name and value the DNA sequence. Although two given species do not necessarily have the same genes, any phylogenetic tree is based on the similarity of their common genes. Indeed,  a phylogenetic tree can be computed as follows. Given a set of species, we extract a set of ``core'' genes (genes present everywhere in this set of species). Each core gene is multi-aligned using a toolbox such as e.g. Muscle~\cite{edgar2004muscle}, and then the alignments are concatenated. A mutation matrix between nucleotides is then built and the estimation of the tree corresponds  to finding the most likely tree that, given this model of evolution, is able to explain the mutations found in the alignment. Statistical techniques like bootstrapping can provide support values on the obtained tree, and these supports provide information on the reliable character of the associated branch. In other words, a branch with a large support value will be considered as trustworthy. 

For the same species, the phylogenetic trees inferred by using two different subsets of common genes are not equal in general. At least, it is unlikely that they share the same branch length. More dramatic, some sister relationship between species may be different (the trees do not have the same topology). Finally, the two obtained trees may share the same topology, but with different support values: some weakness may appear in the first tree in branches considered as trustworthy in the second one. If the given set of species have $n$ genes in common (their core genome has $n$ genes), then we can consider $2^n-1$ different phylogenetic trees that show a certain variability in terms of topology and supports. To choose a tree that faithfully represents the phylogenetic relationship between the species at hand, we may consider that: (1) it is as well supported as possible, and (2) it is based on a nucleotidic sequence which is as large as possible. If such an approach is chosen, our objective is now to find the largest subset of core genes that leads to the tree that receives the largest support. This can be achieved as follows.

The $n$ core genes are sorted alphabetically, and to each subset we associate a binary word of length $n$: its $i$-th character is 1 if and only if the $i$-th core gene is in the considered subset. The alphabet $\mathcal{A}$ is then equal to $\{0,1\}$. Given a binary word $w$, after alignment, we can use a tool like RAxML~\cite{stamatakis2005raxml} to infer a tree as well as its supports. The score $s(w)$ of this binary word can be the average value of the percentage of 1's inside $w$ (number of considered core genes) and the lowest support within the tree (an integer ranging between 1 and 100). This provides us with a score we will have to maximise. We can also consider the inverse of this score, so as to turn the optimisation problem into a minimisation one.

\subsection{A second example: self-avoiding walks and protein folding}

Nowadays, it is relatively easy to have access to the genome of a given species, and to predict its coding sequences, either by blasting~\cite{blast} a set of known genes (basis of knowledge) on it or by using \textit{ad hoc} prediction tools like Glimmer for bacterias~\cite{glimmer}. It is thus quite easy to obtain the DNA sequence of a gene. A more difficult task is to predict its 3 dimensional shape, since this latter is mostly determined by the physico-chemical properties of its amino acid sequence, their coupling interactions, and so on. The protein prediction tools for conformation usually try to find the best conformation according to an optimization function. To do so, they focus on the hydrophylic/hydrophobic properties of each amino acid, and they try to find the 3D conformation that optimizes the number of hydrophobic a.a. inside the conformation (as proteins mainly evolve in aqueous solutions like cytoplasms).

In order to solve this optimization problem, a lattice is first considered, like the 2D or 3D square one, and then, a protein conformation in this lattice will be a self-avoiding walk (SAW~\cite{Dill1985}), i.e. a succession of adjacent nodes of the lattice connected with edges, and such that each node strictly inside the walk has a degree of 2 (as two a.a. cannot lie in the same location), see Figures~\ref{exPliage} and \ref{exElongation}. Given its sequence of hydrophobicity, finding the best 2D conformation of a protein is not an easy task. When considering the set of self-avoiding walks having $n-$steps and whose vertices are either black (hydrophobic) or white squares (hydrophylic residues), the authors of~\cite{Crescenzi98} have indeed proven that determining the SAWs that maximize the number of neighboring black squares in this set, which is supposed to be the conformation chosen by the protein, is NP-hard.

Given a sequence of amino acids, we will resort to using relevant heuristics for predicting the most probable conformation of the protein. The translation in terms of $\mathfrak{S}_n$ and scoring function of Section~\ref{subsec:main} is obvious. Each self-avoiding walk is a word on the \{North, East, South, West\} alphabet, which describes its absolute encoding. The scoring function, here, is the number of neighboring black squares.

\subsection{A third example regarding the protein folding process}
\label{sec:2ndFoldingExample}

\begin{figure}
\centering
\includegraphics[scale=0.6]{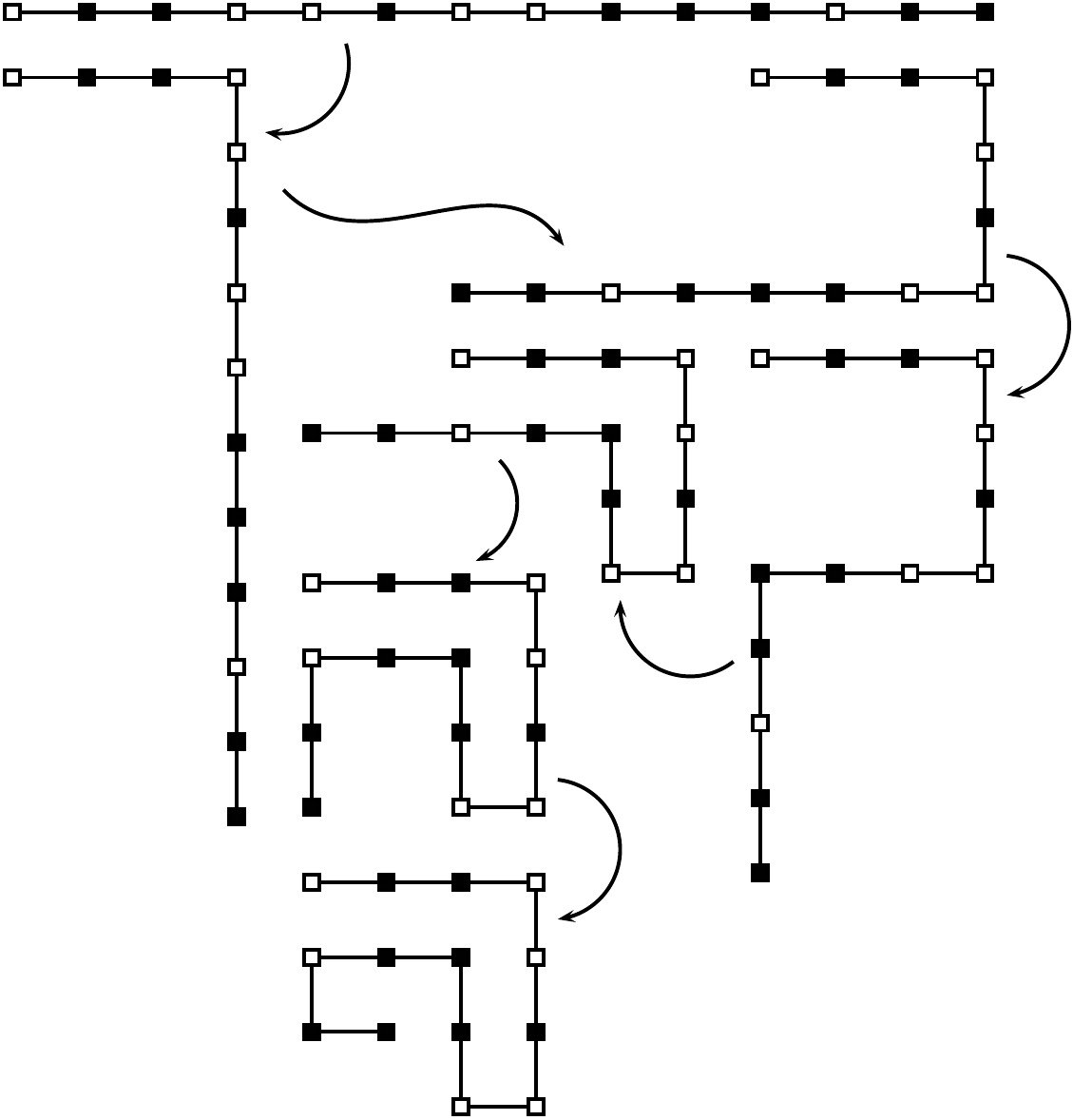}
\caption{Protein Structure Prediction by folding SAWs}
\label{exPliage}
\end{figure}

The problem described in the previous subsection also leads naturally to the study of various subsets of self-avoiding walks~\cite{bgcs11:ij,bgc11:ip,guyeux:hal-00795127}.
In the first approach, starting from the straight line, we obtain by a succession of pivot moves of $\pm 90^{\circ}$ (to choose a node as center of a rotation of the tail) a final conformation being a
self-avoiding walk. All the intermediate conformations must be self-avoiding too (see Fig.~\ref{exPliage}).  
Such a procedure is one of the two most standard implementations of the so-called ``SAW requirement'' in the bioinformatics literature, which is for instance presented in~\cite{DBLP:conf/cec/IslamC10, Unger93,DBLP:conf/cec/HiggsSHS10}. We have previously shown in~\cite{guyeux:hal-00795127} that, \emph{in this first category of prediction software, it is impossible to reach all the possible conformations}.

\begin{figure}
\centering
\includegraphics[scale=0.6]{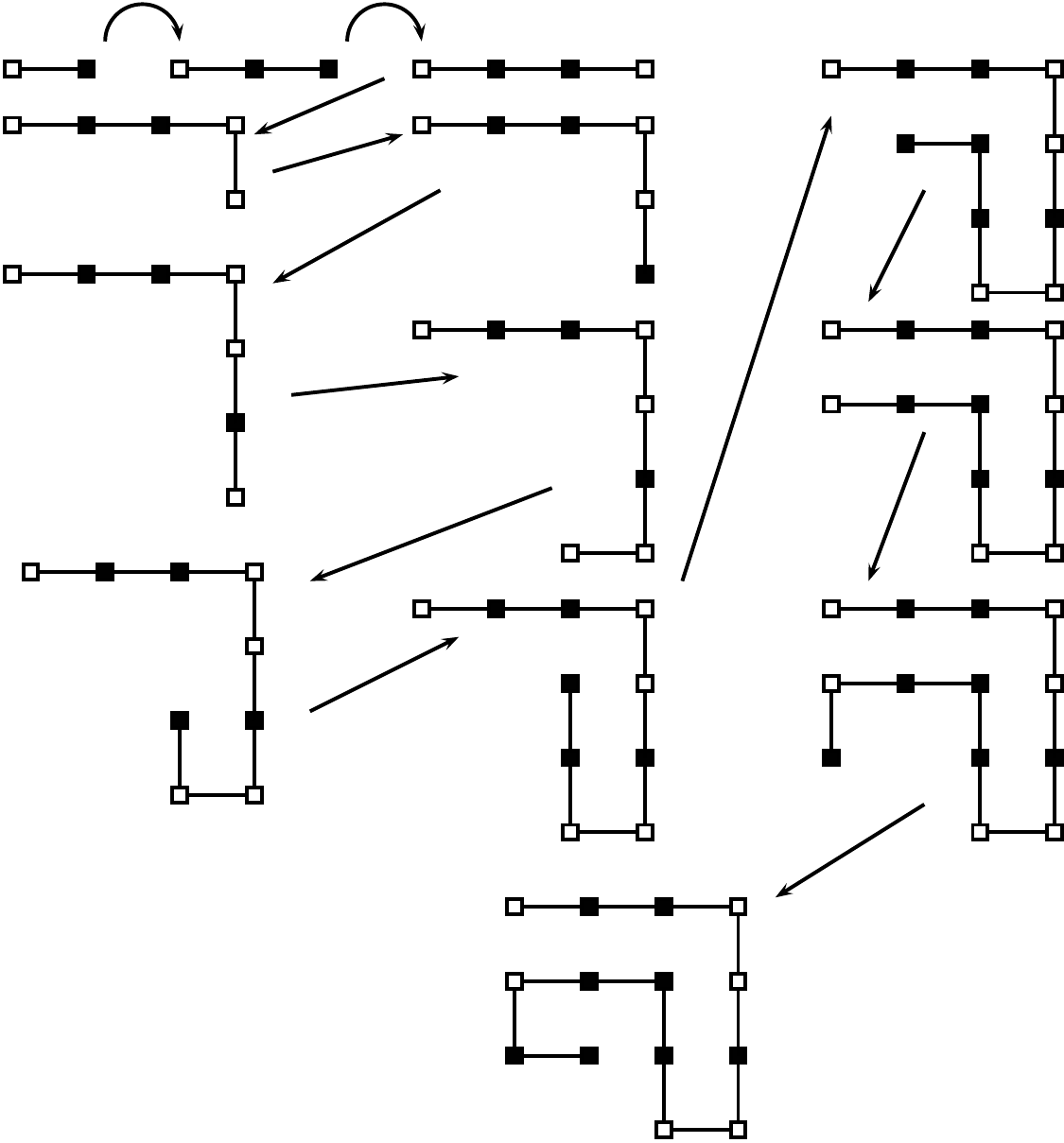}
\caption{Protein Structure Prediction by stretching SAWs}
\label{exElongation}
\end{figure}

The other main approach of protein structure prediction using self-avoiding walks starts with an $1-$step SAW, and at iteration $k$, a new branch is appended to the current tail of the walk, in such a way that the new $k-$step self-avoiding walk has a larger number of neighboring black squares (see Fig~\ref{exElongation}). The protein is thus constructed step by step, reaching the best local conformation at each iteration. It is easy to see that such an approach leads to all the possible self-avoiding walks having the length of the
considered protein~\cite{guyeux:hal-00795127}.

\begin{figure}
\centering
\includegraphics[scale=0.25]{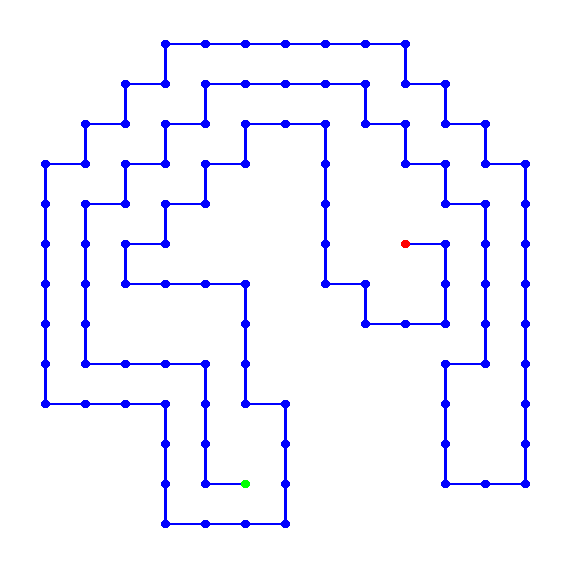}
\caption{An unfoldable self-avoiding walk}
\label{108}
\end{figure}

As stated above, these two approaches cannot reach the same set of conformations. One very bad problem faced by the community is that the difference between the two sets is far from negligible, leading to the necessity of comparing their respective sizes. Clearly, non unfoldable self-avoiding walks, that is, SAWs on which any pivot move introduces an intersection between the head and the tail of the new walk (like in Fig.~\ref{108}), cannot be reached as the result of performing iterative pivot moves starting with the straight line. Our goal is to generate unfoldable self-avoiding walks. In order to do this, a relevant score will be the number of possible pivot moves, which must reduce to zero in the case of an unfoldable self-avoiding walk.

\section{Three embedded heuristics}
\label{sec:3heuristics}

In the present section, we outline the three heuristics that we have embedded within our toolbox. Further details can be found in~\cite{genetic2015,aagp+15:ip} or in the software documentation\footnote{The software is currently downloadable on demand at \url{https://bitbucket.org/rjpGarnier/projetm2recuit.git}}.

\subsection{Genetic algorithm approach}
The population is initialized with a collection of 50 random words. Then, the genetic algorithm iterates until discovering a word whose score is larger than a threshold, or at most for 200~iterations. Each iteration, which produces a new population, consists of the following steps: 
\begin{enumerate}
\item Repeat 5~times a random pickup of a pair $(w^1,w^2)$ of words and mix them using a crossover approach. In this step, indexes $\{1,\dots,n\}$ are partitioned into $k$, $k \le \frac{n}{2}$, subsets $I_1, \dots I_k$. A new word $w$ is then defined by $w_i = w_i^1$ if $i$ belongs to some $I_j$ where $j$ is odd; otherwise $w_i = w_i^2$. The obtained words are added to the population $P$, resulting in population $P_c$.
\item Mutate 5 words of the population $P_c$. More precisely, for each of these words $w$, $k$ randomly selected values of $w$ are replaced by a character randomly picked in $\mathcal{A}$, leading to a new word. The mutated words are added to $P_c$ leading to population $P_m$.
\item Produce population~$P_r$ by adding 5 new random words to $P_m$.
\item Select the 50 best words in population $P_r$ to form the new population~$P$.
\end{enumerate}
The whole set of produced words with their associated scores contains valuable information about which parts of the words tend to decrease the scores. A Lasso test~\cite{tibshirani96regression} may then be applied, to rank the word positions according to their impact on the scoring function. In this case, a last genetic algorithm phase is launched on the updated population, in order to mix these candidate words.

\subsection{Particle swarm optimization approach} \label{sec:PSPI}

The particle position is a vector of $N$ values belonging in $\mathcal{A}$, an alphabet of positive numbers. The objective is to define a way to move the particles in the $N$ dimensional search space so that they produce the optimal vectors with respect to the scoring function.
A swarm of $L$ particles is then a list of position vectors $\left(X_1, X_2, \dots, X_L\right)$ together with their associated velocities $(V_1, V_2, ..., V_L)$, which are $n$-dimensional vectors of real numbers between 0 and 1. The latter are initialized randomly. 
At each iteration, a new velocity vector is computed as follows:
\begin{eqnarray}\label{eq:2}
V_i(t+1)= w V_i(t)+\phi_1\left(P_{i}^{best}-X_{i}\right)+\phi_2\left(P_{g}^{best}-X_{i}\right),
\end{eqnarray}
where $w$, $\phi_1$, and $\phi_2$ are weighted parameters setting the level of each three trends for the particle, which are respectively: to continue in its adventurous direction, to move in the direction of its own best position $P_{i}^{best}$, or to follow the gregarious instinct to the global best known solution $P_{g}^{best}$. Both $P_{i}^{best}$ and $P_{g}^{best}$ are computed according to the scoring function. The new position of the particle is then obtained, by applying the sigmoid function~\cite{intechopen} on each velocity, and 
by considering the value of $\mathcal{A}$ the most close to this latter.

\subsection{Simulated annealing approach}

\begin{figure}
	\centering
	\subfigure[Generic threshold class algorithm]{\centering\includegraphics[width=.55\linewidth]{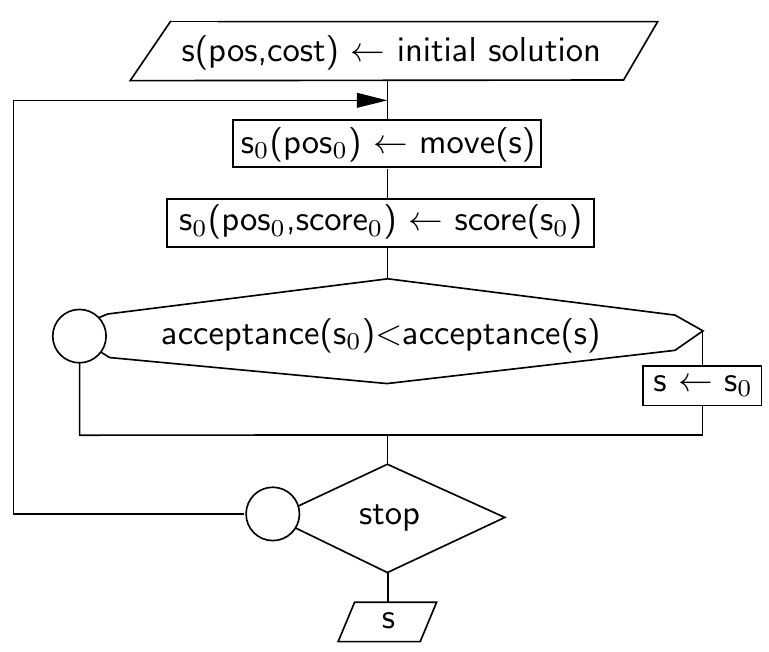}\label{fig:SAClassAlgorithm1a}}
	\subfigure[Metropolis algorithm]{\centering\includegraphics[width=.55\linewidth]{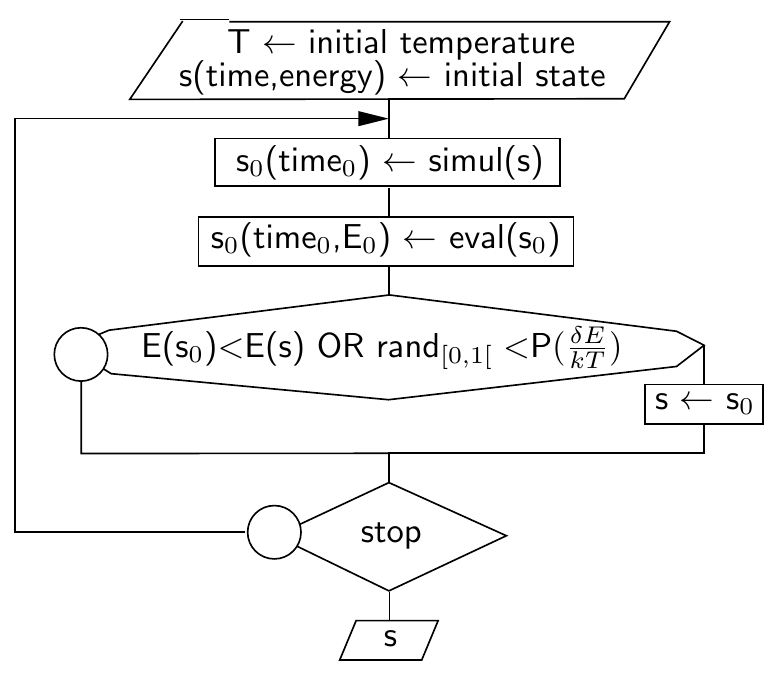}\label{fig:SAClassAlgorithm1b}}
	\subfigure[Simulated annealing algorithm]{\centering\includegraphics[width=.55\linewidth]{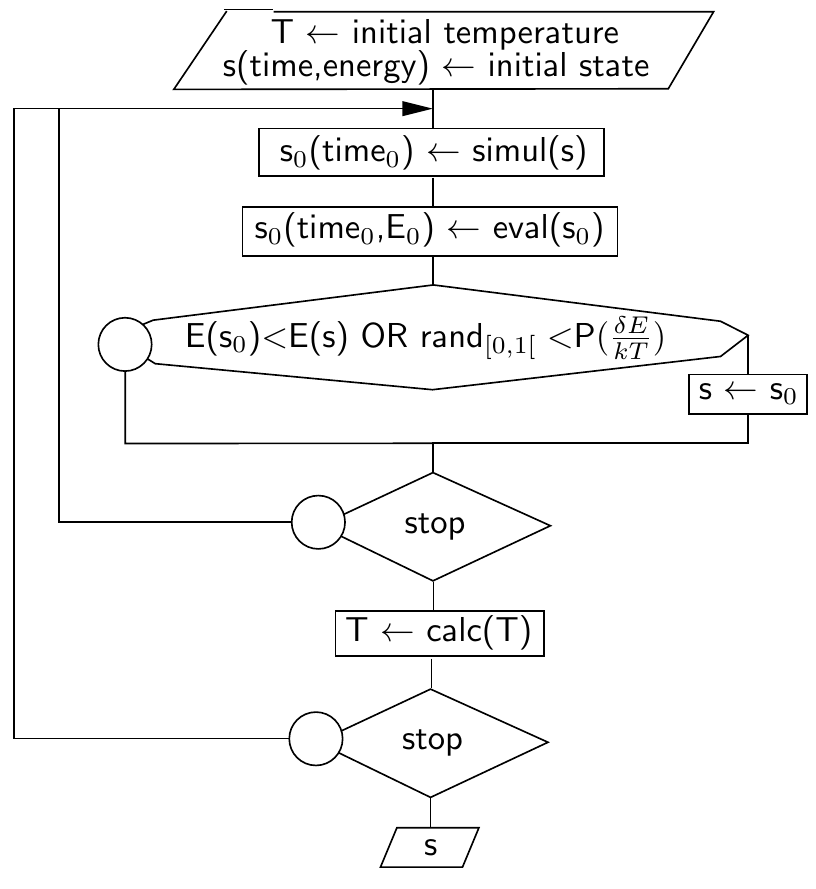}\label{fig:SAClassAlgorithm1c}}
	\caption{Simulated annealing as a threshold class algorithm.}
	\label{fig:SAClassAlgorithm}
\end{figure}

After an initialization step, the Simulated Annealing loop is composed by (a) a move in the neighborhood of the current solution, (b) an evaluation of this new position by a real-valued scoring function, then (c) a test, given a well chosen criterion, to store this position as the new best one. Various criteria can be considered, they all have been embedded within the software.
A detailed description of the simulated annealing algorithm is provided in Figure~\ref{fig:SAClassAlgorithm}.

We are now able to describe the software that embeds these metaheuristics, in order to solve the general problem of Section~\ref{subsec:main}.

\section{Software description}
\label{sec:software}

\subsection{Requirement Guidelines}
Let us first outline the requirements that were set up before the conception of our toolbox.

\begin{itemize}
    \item We need to assess the efficiency of the meta-heuristics (MH) for various models of solution spaces. Therefore, we focus on  measuring and keep track of the performances,  record events or messages (logfiles) related to our algorithms. In other words, introspection is preferred to speed.

\item We need to allow for easy application to new problems the users might come up with. Similarly, new algorithms and heuristics must be easy to embed, as well as batteries of tests. 

\item As we intend to work on multi-cores or clusters and with scoring functions that putatively need a long computation time, we look for a scalable, ready, and reliable network solution. Furthermore, as the heuristic methods under consideration need various amount of resources during computation, we need a convenient scalability mechanism. For this purpose, we chose the Message Passing Interface (MPI) 2.2.

\item Various existing scripts and third party applications need to be embedded or reused without code duplication, and both must be launchable for testing purposes or specific needs. 
Therefore, we need an easy to use but reliable shell script launcher, with basic communication and logging capabilities.

\item Finally, targeted end-users like bioinformaticians are typically familiar with the use of terminals and code scripting, but they are not necessarily computer scientists. This is why the Python language has been chosen. But with a C-like flavor syntax when possible, to produce codes easy to read for everyone.
\end{itemize}

\subsection{Architecture Guideline}	
\begin{figure}
\centering
\includegraphics[scale=.32]{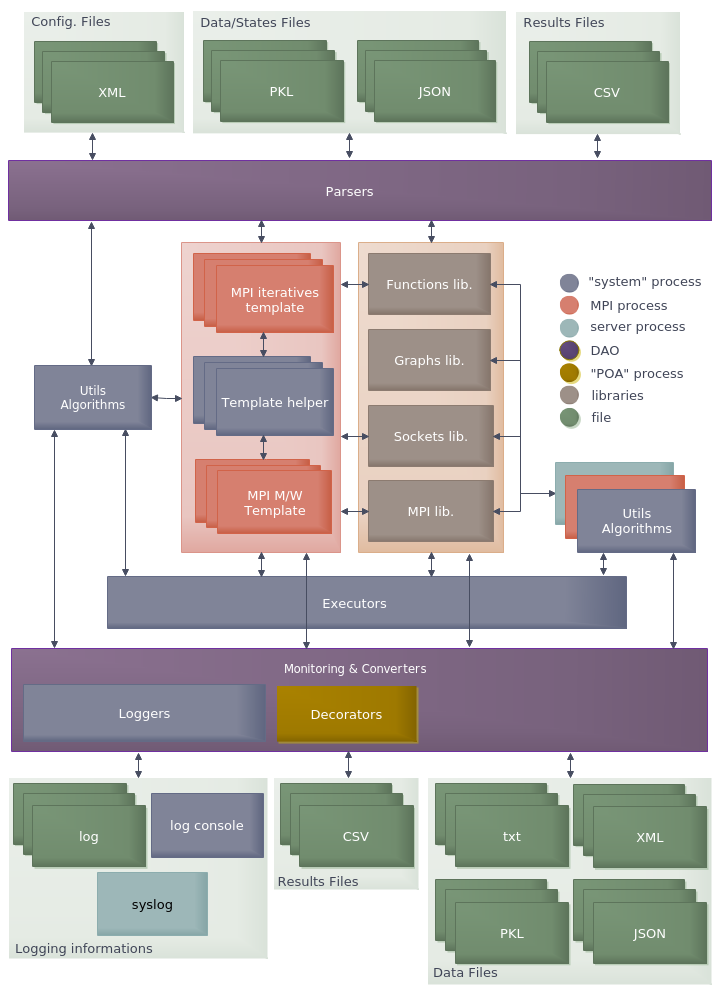}
\caption{Detailed architecture}
\label{fig:GeneralArchitecture}
\end{figure}

Software architecture is meant to convert requirements in a small list of well-known design patterns.
Objectives of the latter is to provide a coherent approach that allows to easily reuse, maintain, and upgrade the proposal. Indeed, a pattern is used whenever possible, while they are reused as soon as possible to reduce their number, leading to an architecture of low complexity as depicted in Fig.~\ref{fig:GeneralArchitecture}. It is described hereafter.

\subsubsection{Introspection -- Decorators}
Oriented Object Programming (OOP) is the main chosen paradigm of the framework but, as Python offers convenient ways to use it, a kind of Aspect Oriented Programming (AOP) has been preferred to separate some technical functions from domain ones. 
Python decorator (wrapper) has been the chosen pattern, which is in this language a simple decoration on a class, function, or method.
Basically a decorator provides a way to intercept communication to or from a designated function. So inputs, outputs, and capability of a decorated class could be analyzed (loggers), validated (contracts), have methods replaced (monkey patching) or enhanced without modifying its code.
Without the necessity to launch tests again on the decorated domain specific code.
The function domain is also more readable and simpler to write. 
Note that a decorator can be shared with as much domain functions as needed. So decorators maximise the reuse of codes, leading to a decrease of both the testing time and the failure risk.

\subsubsection{Framework -- low coupling, helpers, and DAO}

\paragraph{Coupling and cohesion, Liskov principle}
Coupling property is a quality metric that describes the interdependence degree between software modules. Cohesion, for its part, is another metric that describes how much elements of a given module are related on.
Our objective was to increase as much as possible these two metrics, for each module within the software. 
As all MJs use both \emph{move} and \emph{score} concepts (we called them ``field functions''), separated libraries have been built for these functions. Indeed, the generalization of a concrete \emph{move} function leads to a common interface with all other \emph{move} functions. Some characteristics are also shared by all \emph{score} functions, which leads to a second interface. So, each member of our class of functions respects the so-called Liskov substitution principle: if \emph{S} is a subtype of \emph{T}, then objects of type \emph{T} can be replaced with objects of type \emph{S}.

\paragraph{Factory, generic programming, and mapping}

If we have substitutive functions but call a
specific type of them in the main MH template, we need to change code of template each time we change experiment conditions. The coupling must be lower.
We could call a function that is interface compatible. So we need a tool to select
function in a cohesive collection from a XML hashtables and return instance of
this function. This is provided by the factory and dependency injection pattern.

\begin{figure}
\centering
\includegraphics[scale=.50]{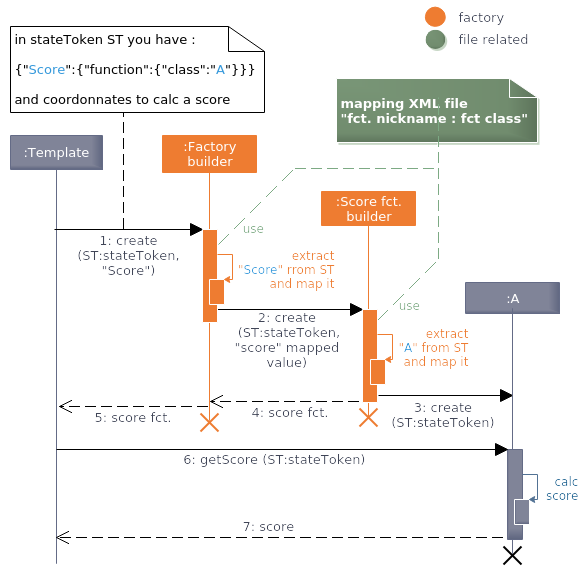}
\caption{Dependency Injection, Factory and Mapping Example (slightly simplified)}
\label{fig:DependencyInjectionSequence}
\end{figure}

As no reference to effective calculus function remains in the main MH code, we abusively speak about \emph{templates}, in a C++ terminology.
Launcher scripts are mostly strategy patterns, \textit{i.e.}, they could be launched by any elemnts of the templates. By doing so, the targeted quality metric was genericity: it measures the ability for effective types of data to be defined dynamically, \textit{i.e.}, only when instantiated.
With this low-decoupling approach, the framework allows a zero-modification reuse of old field functions with a new template, and \textit{vice versa}.

\paragraph{Template helpers}
MH templates often share same concepts but with specific implementation. 
Associated functions of common concept are outsourced.
This approach has been chosen as we wanted not to allow any lambda user to practice a top-down programming. Furthermore, the objective was to discourage a ``scripting'' bottom up style for templates, as authors considered that users will want to plan neither unit nor integration tests on templates. By doing so, code review is more easy to achieve.

\paragraph{DAO and loggers}
As stated previously, various formerly existing shell scripts were at the beginning of this software, and we tried to group them in a unique benchmarkable and reliable environment.
Scripts usually pipe applications, while inputs and outputs of these applications are usually files. 
To be readable, all these files must follow specific standards. 
As a consequence, conversions, sorting, or subset selection are sometimes required regarding these files.
To prevent from issues that may be raised by such a variety of inputs and outputs, two priorities have been kept in mind: an improved and stringent logging and a file system (Data Access Object, \textit{i.e.}, DAO pattern) helpers.

\subsubsection{Maintainability -- Third-party libraries policy}
Authors of this software have \textit{a priori} considered that Python is not the most used language, and so developers may not be familiar with its ecosystem. This lack of background does not allow all developers to sort available libraries according to their maturity or their reliability, while various libraries from the Python ecosystem are still in an early reversion designed from scratch. Consequently, we decided to use as few dependencies as possible, leading to a very simple proposal easy to improve, customize, and fork.

\subsubsection{Scalability -- MPI 2.2 and sockets libraries, server MPI factory}
Let us firstly recall that Message Passing Interface (MPI) is a concurrency inter-process communication system based on message-passing.
The message-passing paradigm implies than process code execution is message-driven, \textit{i.e.}, the type and content of message select the code to invoke for its receiver.
MPI is a Language Independent Specifications (LIS), point to point, and multicast communication protocol. 
It mainly provides a set of basic tools for synchronisation and data exchange, being an ISO 5th, 6th, and 7th level middleware and application framework using TCP sockets. 
APIs are provided for C, C++, and Fortran languages so, in Python, a binding library or a MPI standard implementation is needed. 
This latter is described hereafter.

\paragraph{MPI4py and OpenMPI}
MPI for Python (MPI4py) is based on the standard MPI 2.2 C++ bindings and it provides a similar interface. 
But MPI4py adds an unique Python capability, 
pickling, to send/receive objects in RAM or HDD
buffers.
The widely used middleware OpenMPI 2 has been chosen, as this is a software of production quality which is fully MPI 2.2 and 3.1 compliant. 
It allows process spawning, network and process fault tolerance, safe threads, a good portability, and so on. Last, but not least, it is well documented and supported.

\paragraph{Client/server mode and libraries}
MPI4py over OpenMPI third party library was a prerequisite as most of existing scripts use it.
However, after an intensive evaluation, its low reliably with long series of connections, spawning, or big message exchanges was quite disappointing.
Unfortunately, no good alternative has been found.

As a consequence, we decided to focus on the MPI 2.2 client/server (CS) mode, while minimizing as possible the coupling between MPI4py and the other codes within the software. Indeed, only a MPI-helper library requires it. 

\paragraph{MPI process server factory}

\begin{figure*}[ht!]
\centering
\includegraphics[scale=.4]{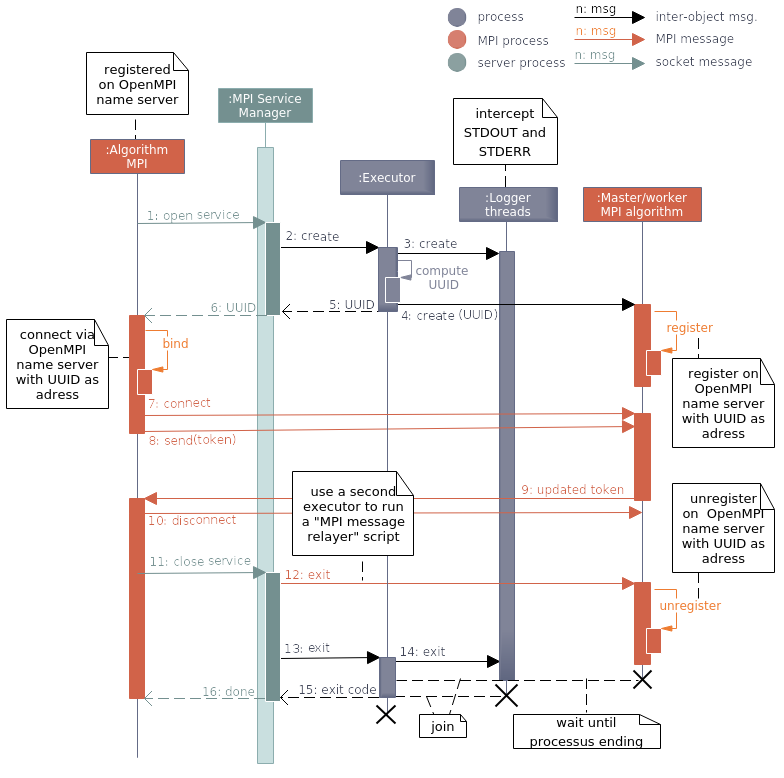}
\caption{Multi-level MPI 2.2 with Server MPI Factory and Executor Class functioning Principles  (slightly simplified)}
\label{fig:MultiLevelMPI}
\end{figure*}

Using OpenMPI name server, a socket helper
library and the OpenMPI helper, we allow a MPI process to launch another MPI
2.2 subprocess (Fig.~\ref{fig:MultiLevelMPI}). And to free resources after results return. This server
XML configured factory with socket communication channel is a standalone
script (singleton).

\paragraph{MPI most used pattern -- Master and Workers}
In this parallel computation pattern, a main thread called the master splits a given problem and merges partial results effectively computed by a set of forked processes, which are called the workers. 
This pattern can rely on a message-passing architecture to broadcast data to workers and regroup partial obtained results.
This is typically a good choice when a problem can be sliced in a set of same smaller and simpler ones, splitting and merging times being small when compared with sub-problems computation ones.

\subsubsection{Reuse with low-coupling -- Executor library}
The MPI server factory needs to execute MPI and Python codes in a shell, and with a maximum of security. 
As a set of applications and scripts has been added that use XML configuration files for their environment, a shell launcher has been required, which should follow an ``easy to use'' philosophy like in Java 5 executor with easy contracting and centralized logging abilities.


\subsubsection{Introspection -- general architecture and dataflow programming}
According to the observation granularity, the proposed framework uses various classical software architectures. 
Some components as the MPI factory server induce their own main architecture, and the classical architecture erosion (gap between models and implementation) is observed. 
This is why, as much as possible, Separation of Concerns (SoC) is used at high level structures.

But a significant choice has been forced and preserved over versions concerning the ``state'' data of each MH process. 
Each problem (that is, the combination of a MH script, field functions like the move or score ones, and their parameters) is defined in an unique XML file.
This XML is validated by XSD; it is parsed, and then converted in a Python dictionary, \textit{i.e.}, a hash-table implementation. 
So (almost) all the needs start in an unique entity. 
The choice has been made to preserve this entity, and to add in it all what MH templates require to be instantiated.
In other words, this is principally a memory for current coordinates and score when moving on the solution space.
This state token, which is a dataflow programming-inspired architecture, is passed to each function and all of them return new version of the token.

With this approach, MH templates can be modeled by graphs, thus simplifying both software monitoring and tuning. 
Additionally, the check-pointing strategy that corresponds to store this token leads to various interesting properties, like the possibility to restore a computation from a given saved point, to study the convergence speed, to benchmark various physical platforms on a same computation, or to launch in parallel various instances to maximize the chance to find interesting solutions (or to study the dispersion of solutions).
We thus improve both introspection and reliability. 

\begin{figure*}[ht!]
\centering
\includegraphics[scale=.35]{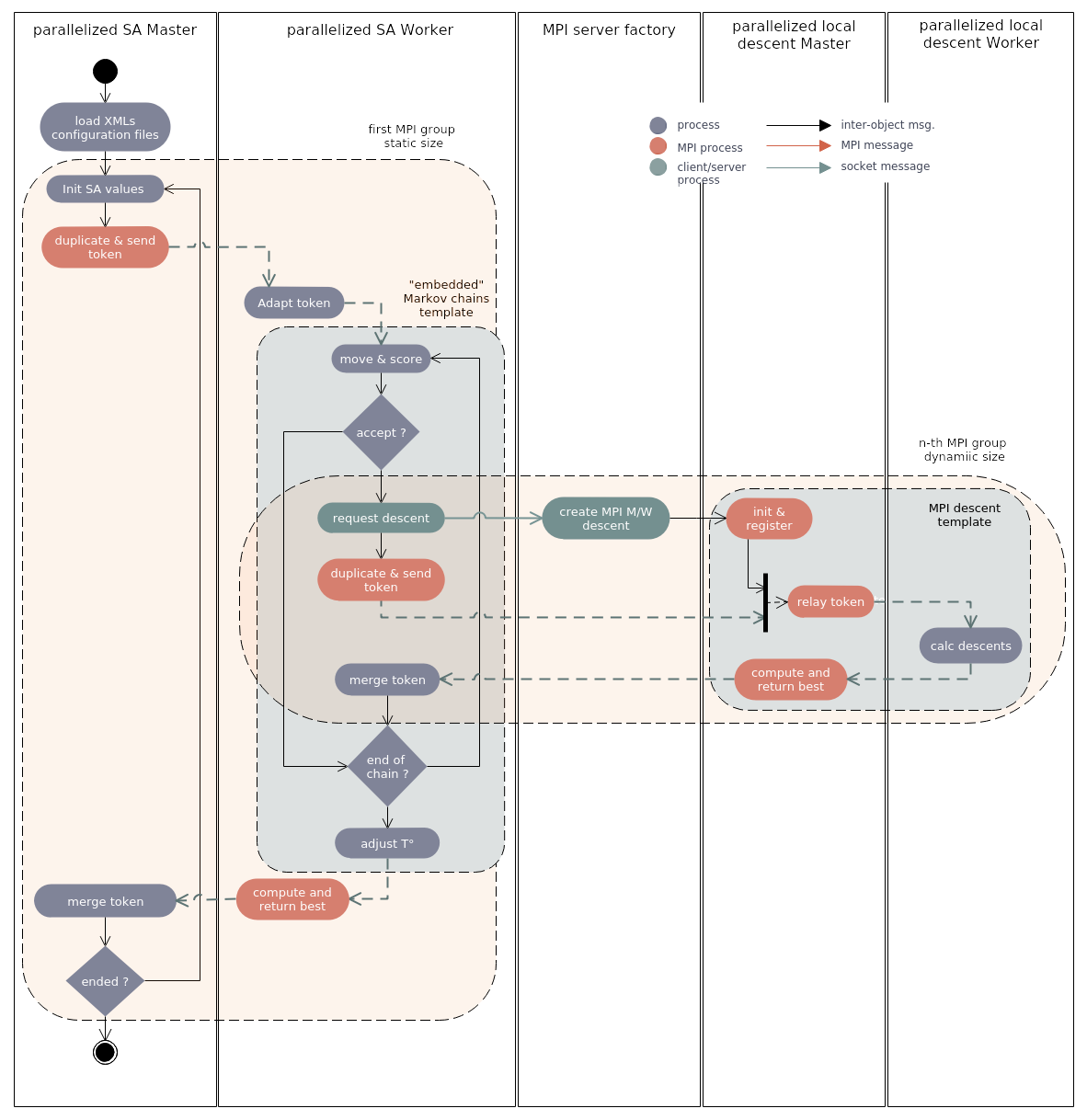}
\caption{2-level MPI 2.2 application : SA with Basin Hopping algorithm (simplified)}
\label{fig:BH}
\end{figure*}

\subsection{How to construct a MH in practice} 

As a framework, a toolbox for computing resources is provided, together with a
methodology to create MH. In broad outline some simple templates are created,
helper are written with DOA and helpers, then when needed, MPI processes are
nested with MPI helper and standalone MPI C/S factory.
Figure \ref{fig:BH} illustrates a SA evolution build by iterating SA and greedy local descent basic algorithms.
The first step was to construct both SA and local descent in an one-core version. 
They are then linked by a constructor to apply a descent on each solution selected by the SA Metropolis criterion. In a second step, the descent is replaced by a MPI version: some descents are launched in Workers and the Master returns the best Worker result. 
The simple Executor is replaced by a call to the MPI server factory. 
Then, in a third step, the simple SA execution itself is injected in a Worker and a Master, created to make a synchronisation after each Markov chain.

On Figure~\ref{fig:BH}, dashed arrows represent data circulation, and then moves of the token instances.

\section{Applications}
\label{sec:appli}

\subsection{Protein folding}
The first illustrative example of the effectiveness of the proposed software is related to the protein folding problem presented in Section~\ref{sec:2ndFoldingExample}. More specifically, given a sequence length, the objective is to find a self-avoiding walk of this length which is non unfoldable: to apply a pivot move at each internal node always leads to a new walk that does not satisfy the self-avoiding requirement.

\begin{figure}
    \centering
    \includegraphics[scale=0.09]{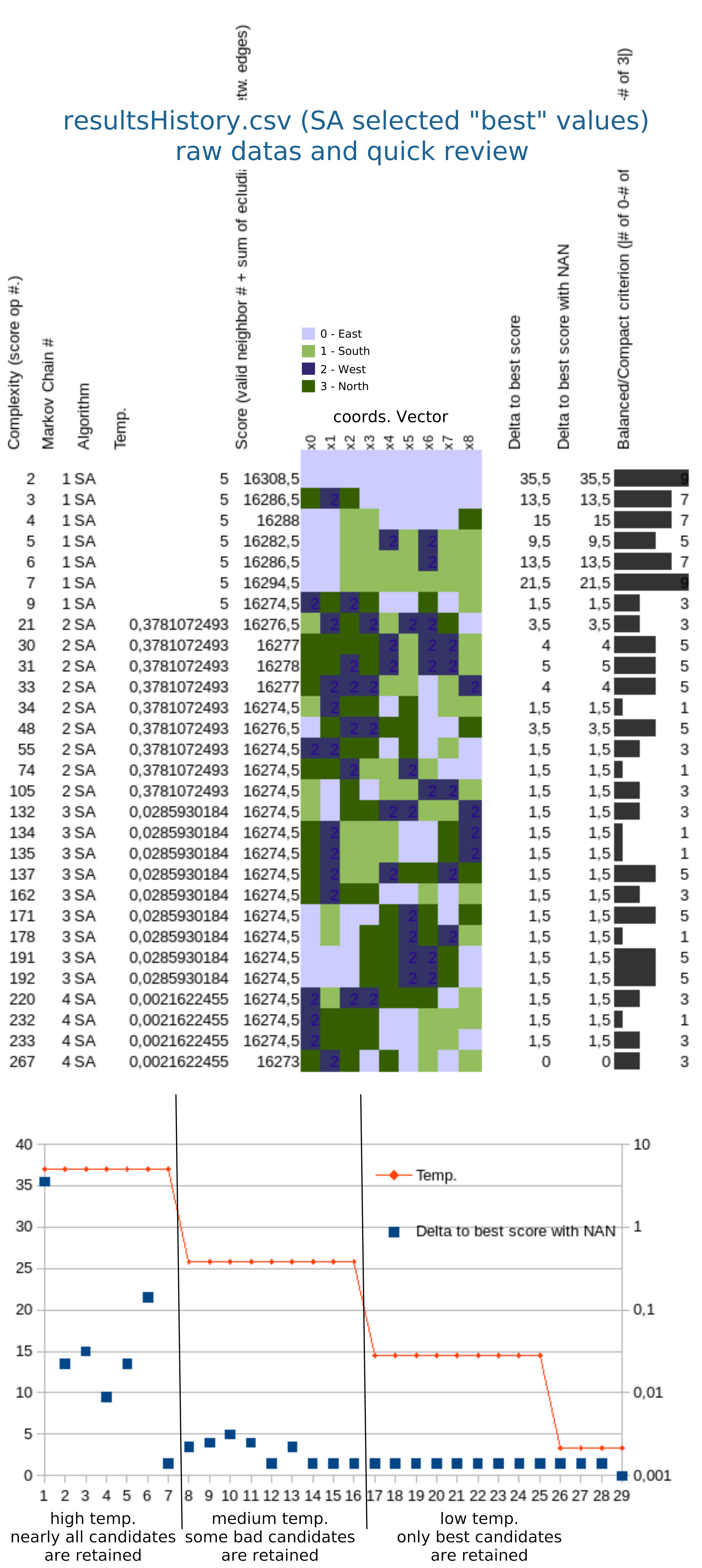}
    \caption{Example of output exploitation using the SA on a short size problem.}
    \label{fig:SAprotein}
\end{figure}

We considered the number of nodes on which a pivot move can be applied as the scoring function. When this score is zero, then a non unfoldable self-avoiding walk has been obtained. As a first basic experiment, we considered the SA metaheuristic on short sequences, to reduce the computation time. Obtained results are depicted in Figure~\ref{fig:SAprotein}, which emphasize the convergence ability of the proposed software on the considered problem.

\subsection{Phylogeny}

\begin{table}
\centering
\begin{tabular}{c|l|c}
Accession Nb & Name & Nb. of genes\\
\hline
NC\_024082.1 & Cylindrotheca closterium & 257\\
NC\_014808.1 & Thalassiosira oceanica CCMP1005 & 138 \\
NC\_025313.1 & Cerataulina daemon & 195 \\
NC\_028052.1 & Pelargonium cotyledonis & 271\\
NC\_015403.1 & Fistulifera solaris & 192  \\
NC\_024084.1 & Leptocylindrus danicus & 155  \\
\end{tabular}
\caption{Family number 1 (Pelargonium cotyledonis as outgroup).}
\label{tab:famille13}
\end{table}

\begin{figure}[ht]
    \centering
    \includegraphics[scale=0.4]{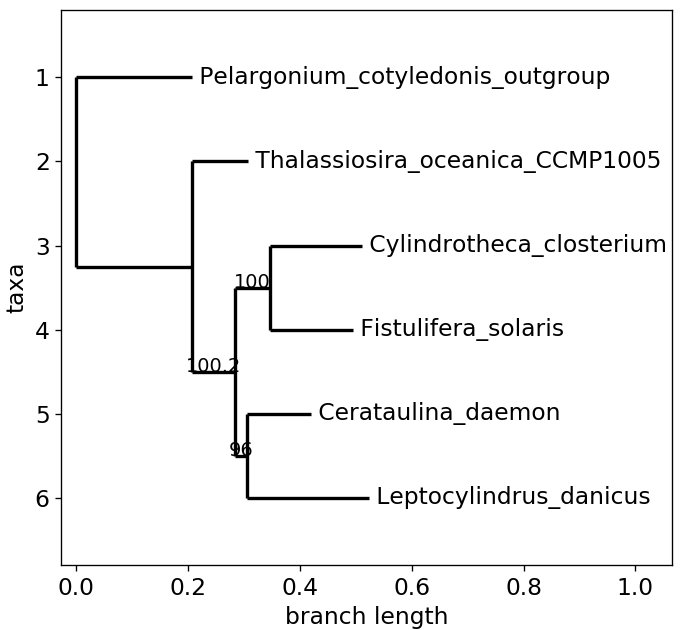}
\label{fig:famille13topo0GA}
    \caption{First obtained supported topology.}\label{fig:famille13topologies}
\end{figure}

We have first considered phylogenetic issues, with the family listed in Table~\ref{tab:famille13}. 
This family is constituted by 6 genomes, with a 
number of detected genes 
that ranges from 138 to 271, and a core genome of 122. The phylogeny with the alignment of these core genes leads to a small weakness in one branch (bootstrap lower than 95). 
%
%
To wonder whether some genes may be responsible of such weak uncertainty, we have firstly launched the genetic algorithm, which 
has stopped after 29 iterations, in systematic investigation mode, leading to 2 topologies:
\begin{itemize}
 \item Topology 0 depicted in Fig.~\ref{fig:famille13topo0GA}, that has occurred 27 times. The best obtained tree has a lowest bootstrap of 96, while in average the lowest bootstrap is equal to 86.
 \item Topology 1, for its part, has occurred twice, with a non supported branch of 64 in its best tree. It is the same than Top.~0, except that the clade in newick format ((\textit{C.closterium},\textit{F.solaris}),(\textit{C.daemon},\textit{L.danicus})) becomes (((\textit{C.closterium},\textit{F.solaris}),\textit{C.daemon}),\textit{L.danicus}).
\end{itemize}

The PSO, for its part, 
has rapidly found a first well supported phylogenetic tree in a third different topology, and with all supports equal to 100, namely: (((\textit{C.closterium},\textit{F.solaris}),\textit{L.danicus}),\textit{C.daemon}). However, the PSO has used only 47.5\% of the core genes to reach such a tree. According to our stop criterion, this tree has not been returned by the algorithm. Indeed, this example illustrates the ability of the particle swarm optimization algorithm to more globally visit the whole space at the beginning, in order to discover regions of interest. 


The simulated annealing, for its part, raised these 3 topologies. 
A remarkable element is that these 3 topologies have the whole bootstraps equal to 100. Furthermore, Topology~2 appears as the best one according to the produced result (it was Topology~0 according to the GA, while PSO has not succeeded in separating these two topologies). 
Obviously, both PSO and SA have converged to local minima that are not global ones if we consider that both minimum bootstraps and proportion of core genes must be maximized. Launching them again with other initial values and parameters may select other optimal positions in the cube. 

\section{Conclusion}
In this article, we have detailed the general principles at the bases of an efficient software we propose for the bioinformatician community. This software iterates on finite words on a finite alphabet, such that each word has a score. Using one of the three most classical metaheuristics, this software is able to discover the optimums in a distributed manner. The software is easy to configure, to use, and to update. It can resolve a variety of bioinformaticians problems, like to measure the effects of genes on a phylogeny.

In future work, we intend to improve the following elements. First of all, the XML ``problem'' configuration file must have a more common structure, to allow state token to be used by more than one specific MH template. The objective is as follows: a problem, started to be resolved by a SA, should be finalized by a PSO, and \textit{vice versa}. 
An improvement of the execution speed must be achieved, among other things by substituting some of our functions by more efficient and reliable ones provided by third-party libraries. Finally, the number of tests, contracts, and code factoring must be enlarged, together with logs, error management, and documentation, to help bug tracking, maintainability, safety, and data integrity.

\bibliographystyle{unsrt}
\bibliography{biblio}

\end{document}